%% file: neurips_2022.tex
\documentclass{article}
\pdfoutput=1

\usepackage[preprint]{neurips_2022}

\usepackage[utf8]{inputenc} 
\usepackage[T1]{fontenc}    
\usepackage{hyperref}       
\usepackage{url}            
\usepackage{booktabs}       
\usepackage{amsfonts}       
\usepackage{nicefrac}       
\usepackage{microtype}      
\usepackage{xcolor}         
\usepackage{graphicx}
\usepackage{subcaption}

\title{Toward Human-AI Co-creation to Accelerate Material Discovery}

\author{%
  Dmitry Zubarev \\
  IBM Research Almaden \\
  San Jose, CA, USA \\
  \texttt{dmitry.zubarev@ibm.com} \\
  \And
  Carlos Raoni Mendes\\
  IBM Research Brazil \\
  Rio de Janeiro, RJ, Brazil \\
  \texttt{craoni@br.ibm.com} \\
  \And
  Emilio Vital Brazil \\
  IBM Research Brazil \\
  Rio de Janeiro, RJ, Brazil \\
  \texttt{evital@br.ibm.com} \\
  \And
  Renato Cerqueira\\
  IBM Research Brazil \\
  Rio de Janeiro, RJ, Brazil \\
  \texttt{rcerq@br.ibm.com} \\
  \And
  Kristin Schmidt \\
  IBM Research Almaden \\
  San Jose, CA, USA \\
  \texttt{schmidkr@us.ibm.com} \\
  \And
  Vinicius Segura \\
  IBM Research Brazil \\
  São Paulo, SP, Brazil \\
  \texttt{vboas@br.ibm.com} \\
  \And
  Juliana Jansen Ferreira \\
  IBM Research Brazil \\
  Rio de Janeiro, RJ, Brazil \\
  \texttt{jjansen@br.ibm.com} \\
  \And
  Dan Sanders \\
  IBM Research Almaden \\
  San Jose, CA, USA \\
  \texttt{dsand@us.ibm.com} \\
}

\begin{document}

\maketitle

\begin{abstract}
There is an increasing need in our society to achieve faster advances in Science to tackle urgent problems, such as climate changes, environmental hazards, sustainable energy systems, pandemics, among others. In certain domains like chemistry, scientific discovery carries the extra burden of assessing risks of the proposed novel solutions before moving to the experimental stage. 
Despite several recent advances in Machine Learning and AI to address some of these challenges, there is still a gap in technologies to support end-to-end discovery applications, integrating the myriad of available technologies into a coherent, orchestrated, yet flexible discovery process.
Such applications need to handle complex knowledge management at scale, enabling knowledge consumption and production in a timely and efficient way for subject matter experts (SMEs).
Furthermore, the discovery of novel functional materials strongly relies on the development of exploration strategies in the chemical space.
For instance, generative models have gained attention within the scientific community due to their ability to generate enormous volumes of novel molecules across material domains.
These models exhibit extreme creativity that often translates in low viability of the generated candidates. 
In this work, we propose a workbench framework that aims at enabling the human-AI co-creation to reduce the time until the first discovery and the opportunity costs involved.
This framework relies on a knowledge base with domain and process knowledge, and user-interaction components to acquire knowledge and advise the SMEs. Currently, the framework supports four main activities: generative modeling, dataset triage, molecule adjudication, and risk assessment.
\end{abstract}

\section{Introduction}\label{Introduction}
In the technological era, faster scientific discovery grants higher resilience at societal and individual level to the ever-emerging challenges, such as the latest pandemic. Even under normal circumstances, the entities performing discovery tasks, from the level of individual researcher to an academic or industrial organization, face multiple inefficiencies and bottlenecks that hinder overall progress.
To illustrate some of these challenges, we will focus on the discovery of polymeric materials and mitigation strategies rooted in human-AI co-creation although many of the raised issues pertain to other domains of chemical and material sciences~\cite{Pyzer-knapp2022,Suh2020}.

Any chemical product, from a small-molecule drug to a piece of plastic wrapping, has a non-trivial footprint over its lifecycle. During the discovery phase, the assessment of new material cannot be restricted to the functional characteristics anymore. 
It must include a comprehensive set of criteria ranging from the nature of the synthetic pathway (i.e. are the precursors from petrochemistry or bioengineered; are the catalysts problematic; etc.) to the possible degradation pathways in the environment, patterns of bioaccumulation and metabolism, as well as overall toxicity. 
Unfortunately, this was too often an after thought, and technological advancements causes severe consequences for humans and the environment. 
The old pattern of discovery that disregards technological and environmental risks of new materials is not tenable anymore~\cite{Hoffman2021}.
A separate category of risks is associated with the discovery process as such. After all, discovery requires investment of appreciable resources with the goal of achieving such level of success that will offset the costs. 
Here, “resources”, “costs”, and “success” should be understood in the broadest manner.
For example, academic discovery measures success primarily in the volume of published papers as a proof of work, whereas industrial discovery relies on development of intellectual property and reduction to practice. 
In any case, the discovery process is associated with risks, such as inability to solve the task within a given budget or given time, unacceptable opportunity costs, inadequate selection of discovery targets, and so on.

In the contemporary scientific discovery, subject matter experts (SMEs) are the owners and the stakeholders. 
Regardless of comparative performance of SMEs and AI systems in various specialized tasks, these roles will remain uniquely human in the foreseeable future. 
These roles fundamentally dictate the selection of goals and targets: project scoping, development of the technical specifications for the discovery components, establishing constraints, and making actionable decisions fall under SME's responsibility and require adequate level of expertise and knowledge. 
Expert knowledge is conceptual and supports reasoning within the specific scientific domain. 
In a sense, expert knowledge bears similarity to “common sense” knowledge, considering that it must be shared among domain experts as a common denominator enabling scientific discourse. 
Expert knowledge is combined with tacit knowledge, which is highly personalized as a reflection of the unique scientific trajectories of the experts. 
A part of the tacit knowledge is attributable to the institutional knowledge encapsulating concepts and practices of institutions (cf., scientific school of thought). 
The nature of SME's knowledge is such that it cannot be efficiently captured by contemporary AI systems because it is not captured in structured or unstructured sources. 
Simultaneously, SME's knowledge and expertise have glaring deficiencies: they are incomplete, inseparable from humans' cognitive biases, and coupled with the fundamental irrationality of human decision-making. 
Therefore, the role of the SMEs in the discovery process is to contribute complementary cognitive capabilities and conceptual knowledge as the primary agent in the discovery process, while knowingly mitigating counterproductive aspects of human intelligence via interactions with AI components~\cite{Abdul2018,kip}.

In summary, the human-AI interaction is critical in the development of new AI-based approaches and their supporting technologies to accelerate materials discovery.
We advocate for an integrated platform to empower SMEs by leveraging the human-AI co-creation symbiosis to reduce the time to discovery and the opportunity costs.
In this work, we will discuss a knowledge-driven discovery workbench as a platform that integrates four main activities:
(1) generative modeling, (2) dataset triage, (3) molecule adjudication, and (4) risk assessment. 
In the next section, we present the reference use case that has been driving our experiments and will be used in the following sections.
In this use case, we enabled the integration of the four activities to support SMEs in discovering new Photoacid generators (PAG), a critical class of materials used in semiconductor manufacturing, and evaluated candidates more closely for environmental risks~\cite{Hoffman2021}. 
Afterwards, we discuss the four main activities currently supported by the proposed discovery workbench framework.

\input{tex/pag-use-case}

\input{tex/DWb}

\subsection{Generative Modeling} \label{GenerativeModels}
Let’s compare discovery of materials and small-molecules. 
Primary difficulties of small-molecule discovery are large cardinality of the set of molecules (often estimated as $10^60$) and severe limitations of synthetic accessibility of large fraction of this set. 
Discovery of polymer materials includes synthesis of precursors as the initial step, so the large but finite cardinality of the set of molecular precursors, catalysts, and molecular additives is a contributing factor. 
However, polymerization itself, polymer formulation, and material processing require choosing values of continuous degrees of freedom with infinite cardinality. 
Moreover, there’s a separate combinatorial explosion of the experimental protocols transforming a specific precursor into the material suitable for a technological application.

Design of experiments solving search and optimization tasks under such circumstances must be as efficient as possible to meet practical constraints on time and cost of discovery. 
Generative models (GMs) offer an attractive route of finding discovery targets, or at least plausible candidates, by expanding sets of known solutions to include new, generated candidates according to learnt joint probability distribution of the features over the initial set. 
Technical details of various generative modeling strategies can be found in related work like \textit{Hoffman et al.}~\cite{Hoffman2021} and \textit{Maziarka et al.}~\cite{Maziarka2020}. 

Operationally, the main question that needs to be addressed in application of generative models to materials discovery is how to reconcile the volume of the generated structures with finite throughput of chemical processes, regardless of the level of automation and autonomy. 
GMs in practice require a some auxiliary screening and constraining via discriminative models and filters. Tacit expert knowledge plays enormous role here – in lieu of GMs that can blend conceptual knowledge with factoids, SMEs assume discriminative role either directly or indirectly, via models of the SMEs decisions. 
Level of acceptance of GM candidates by SMEs can vary appreciably among different GM models and depending on the nature of SMEs task. 
For example, materials scientists are being more stringent in evaluation of synthetic practicality of new candidates and more accepting in evaluation of their functional potential.

In the end-to-end materials discovery the AI component based on a GM is expected to generate hypotheses that pass peer-review by the SME who is the owner of the discovery task. 
This leads to the first metric of GM quality assessment, the fraction of candidates that pass SME-driven peer-review. 
We would like to emphasize that this is an observed and factually relevant metric, and it does not imply that SME decisions are flawless. 
This metric equally applies "autonomous lab" discovery - any autonomous scientific equipment has an SME owner and stakeholder who is directly responsible for the efficiency and cost of platform utilization. The secondary metrics include comparison of the generated and initial samples in terms of the geometry (topological data analysis on point clouds, manifold learning, and such), and statistics of the constitutional features and properties as proposed by \textit{Tadesse et al.}~\cite{Tadesse2022}. 

\input{tex/dataset-curation}

\input{tex/risk-assessment}

\input{tex/final-remarks}

\bibliographystyle{plain}
\bibliography{references}

\end{document}

%% file: tex/pag-use-case.tex
\section{The PAG use case} \label{PAGUseCase}

Sustained growth of computing platforms is contingent on sustainable development of semiconductor technology, which cannot be taken for granted, as the pandemic-induced chip shortages demonstrated. 
Technology of chemical amplification is the foundation of high performance of semiconductor lithography and photoacid generators (PAGs) are one of the many materials used in the semiconductor manufacturing. 

Unfortunately, PAGs are often toxic and not degradable as many of them contain per- or polyfluoroalkyl  substances (PFAS). PFAS materials are under increasing scrutiny due to their harmful environmental and human health impacts. There is a clear need for bio-degradable, non-toxic, yet efficient new PAG materials. We choose this use case as it encompasses a range of discovery challenges: knowledge is sparse and distributed across different domains (i.e. semiconductor manufacturing, photo catalysis, toxicity), and many different aspects need to be taken into account to assess potential new candidates (e.g. environmental assessment, IP/novelty, efficiency, synthetic viability), which requires SMEs from different fields to collaborative work together on a solution.

\input{fig/pag-data-visualization}
We implemented a process for PAG discovery where AI and different SMEs arrive at a new PAG candidates in a collaborative manner.
This process starts with integrating historical data with generative models, the idea is to extract examples of molecules from documents, e.g., patents, to feed the AI-based molecule generators~\cite{Maziarka2020,Shi2020}. Both the inputs to the generative models and outputs (chemical structures of PAG cations) are computationally characterized to produce labels including wavelength of the highest photon absorption  $\lambda_{max}$, biodegradability, and 50\% lethal dose LD50.
Figure~\ref{fig:pag-data-visualization} shows the historical data along with the result of generative models for the PAGs.
In practice, we processed approximately 5K patents, extracted approximately 4K structures of ionic PAGs comprised of 1.3K cations and 0.5K anions. The sample produced by the generative models comprised approximately 10K PAG cations.
Since the amount of molecules generated is much more than the capacity of synthesizing them, we introduce peer-reviewing mechanisms where SMEs interact with the generative models. First, SMEs assign viability labels to a sub-sample of the generated structures in \textit{adjudication phase} by answering the question "Does this candidate merit further exploration?". 
Here, we do not resolve the factors dictating the SME's decision and focus on learning the general pattern of rejections.
Typically, they are associated with issues of synthetic costs, scale-up possibility of synthesis, possibility of undesirable secondary chemical processes, etc. 
The adjudication phase enables us to model the rejection priorities of the SMEs and prioritize the entire generated sample. Second, SMEs define specialized cutoffs based on the application and according SMEs' specialization, for example, functional application, generation of intellectual property, and environmental regulations. Along with the adjudication labels, these explicit filters enable \textit{triage phase} where we reduce the number of molecules as much as by three orders of magnitude and rank them according to their viability as discovery targets.
Finally, the set of molecule candidates goes through a risk assessment methodology where a multidisciplinary group of SMEs evaluates the molecule to decide whether it will be synthesized or not based on the known-unknowns and known-characteristics (Section~\ref{RiskAssessment}).

%% file: fig/pag-data-visualization.tex
\begin{figure}
    \centering
    \includegraphics[width=0.7\textwidth]{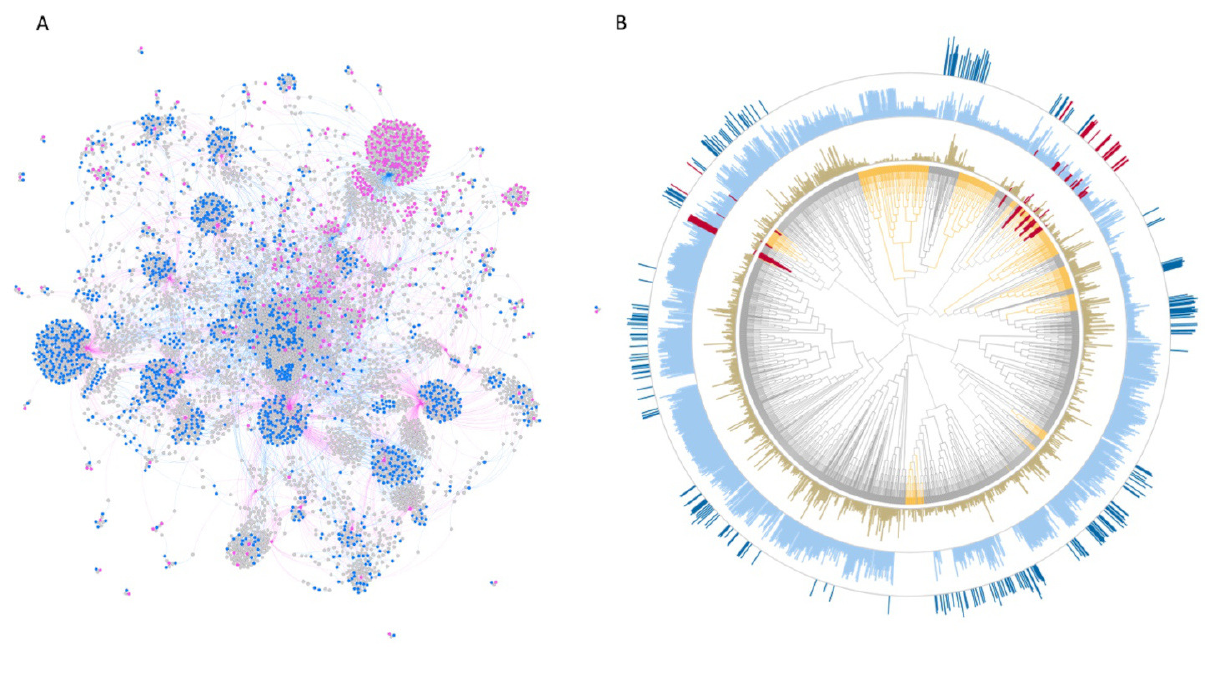}
    \caption{Panel A shows constitutional similarity network of ionic PAGs found in patents. Grey nodes are PAG salts (ionic pairs), pink nodes are anions, and blue nodes are cations. Nodes of the cation and anion comprising the salt are connected to the node representing the salt. Panel B shows phylogenetic tree of PAG cations extracted from patents (grey branches) and produced by generative model (orange and red branches), where red branches represent generated cations accepted by SMEs. Radial bar charts represent computed properties: LD50 (tan), Biodegradability (light blue), and $\lambda_{max}$ (blue).}
    \label{fig:pag-data-visualization}
\end{figure}

%% file: tex/DWb.tex
\section{Discovery Workbench}\label{sec:DWb}
To enable AI-assisted applications that support scenarios like the PAG use case, we created the \textit{Discovery Workbench} (DWb) framework, which aims at accelerating the construction of end-to-end discovery applications. DWb offers mechanisms to integrate multiple discovery tools and to support common discovery workflows by means of an integrated and user-friendly web-based interface.
DWb provides systematization to support experts with the discovery process and tracks the knowledge that is created along that process throughout its entire life cycle, from hypothesis generation and collection (\textit{i.e.} material candidates), to hypothesis testing (\textit{i.e.} automated simulations and lab experiments), and then to drawing conclusions (\textit{i.e.} visualizations and comparisons).






DWb is based on the idea of capturing and organizing the knowledge created and obtained during a discovery process executed by SMEs, so it can be reused in future iterations to accelerate other discovery campaigns.
To achieve this, DWb is build around few core concepts, some of them being customized per domain (figure~\ref{fig:dwb-core-concepts}):
    \textbf{Domain Object} The finest granularity of instances that the system will handle, which is customized per domain and can be anything (\textit{e.g.}, molecules, complex materials, oil reservoirs);
    \textbf{Characterization} A collection of \textit{Attribute Values} for a \textit{Domain Object}, representing a possible representation of that object;
    \textbf{Dataset} A collection of \textit{Characterizations}; 
    \textbf{Study} A focused investigation created by domain experts, with associated \textit{Datasets} of interest in which SMEs can perform \textit{Activities} toward a certain discovery goal;
    \textbf{Activity} A framework plugin represented by a workflow that consumes and produces \textit{Datasets}, refining, filtering, or transforming \textit{Characterizations}, and it may involve \textit{investigative tasks} (in which users interpret data like in molecule adjudication and risk assessment) and/or computational \textit{operations};
    \textbf{Operation} Another framework plugin that executes an algorithm that typically produces a new \textit{Dataset} (\textit{e.g.}, generative models, triage).

\input{fig/dwb-core-concepts}

DWb provides multiple ways for SMEs to interact with the system, empowering them to perform their daily activities.
For example, in the PAG use case, we can follow the execution of a ``triage'' operation with an ``adjudication'' activity or a ``virtual experiment'' one (shown in Figure~\ref{fig:dwb-pag-workflows}). 

\input{fig/dwb-pag-workflows}

From a technical standpoint, DWb is a combination of several components and its architecture focuses on extensibility in multiple levels: adaptation to new domains through extensions to its core ontology, integration of new plugins, and customization of deployments following a Software Product Line approach.
In the following sections, we will present the main activities currently supported by DWb.


%% file: fig/dwb-core-concepts.tex
\begin{figure}
    \centering
    \includegraphics[width=0.7\textwidth]{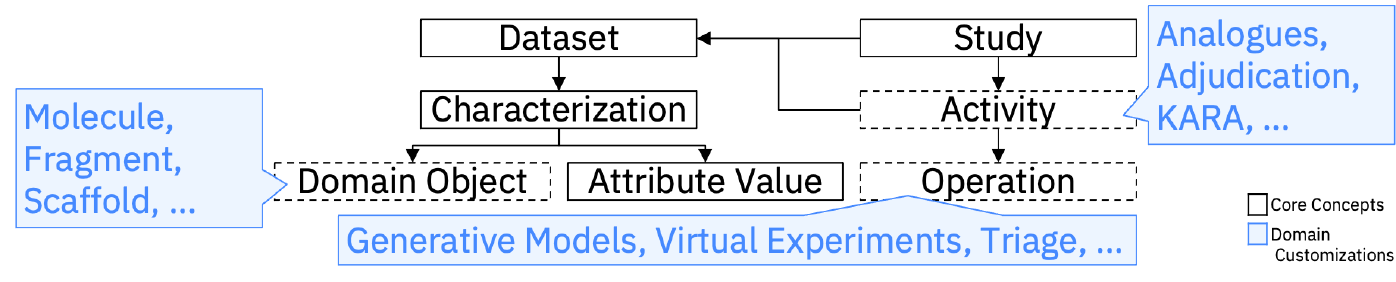}
    \caption{Core concepts of the DWb (in black) and the domain customizations (in blue).}
    \label{fig:dwb-core-concepts}
\end{figure}

%% file: fig/dwb-pag-workflows.tex
\begin{figure}[b]
    \centering
    \includegraphics[width=0.7\textwidth]{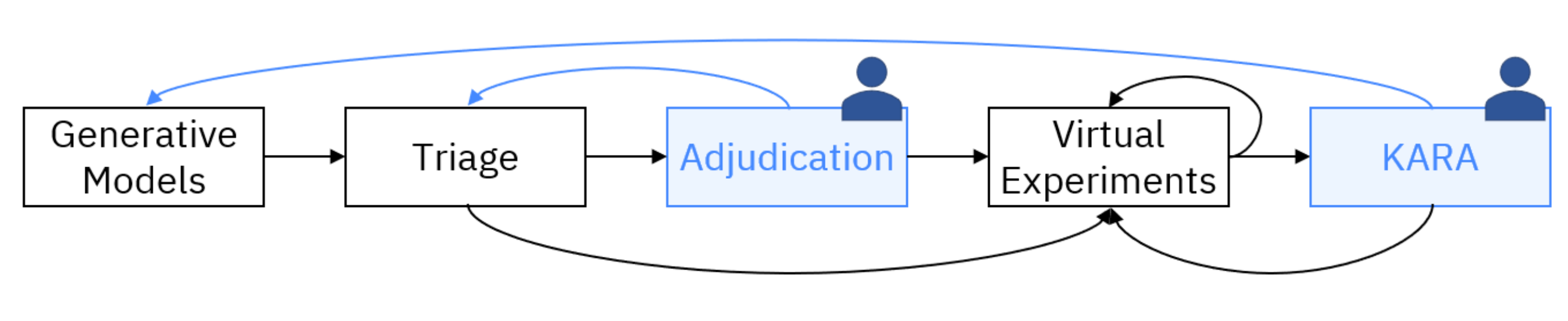}
    \caption{Example of workflows supported by DWb in the PAG use case. Blue boxes show expert-in-the-loop activities and blue arrows highlight machine teaching opportunities.}
    \label{fig:dwb-pag-workflows}
\end{figure}

%% file: tex/dataset-curation.tex
\subsection{Dataset Curation} \label{sec:dataset-curation}
After the generative model execution, the materials' dataset can be very extensive with numerous entries that are unrelated with the current task at hand (in our experience, about $10^5$ candidates).
While this is the generative model strength (quickly generate an enormous number of candidates), it is also its downsize -- it is impractical to investigate each individual candidate.
Thus, we follow the generative model with a \textit{dataset curation} process to select/prioritize the materials.

We based this process in two strategies: one based on the implicit restrictions specific to the user and the study and another based on the computable properties.
There are numerous restrictions that will never become explicit, as it has to do with (i) the uniqueness of the structure and (ii) the knowledge access.
Due to the nature of Generative Model, it can pick up such patterns of atom and functional group arrangements that would not come to the SME's mind -- so they cannot say upfront that some of those arrangements would be unfavorable, therefore not being able to create a restriction for the model.
Moreover, many restrictions are institutional and/or tacit knowledge, not being able to extracted them from literature as they are specific to a group and may even be trade secrets.

Since we have this implicit knowledge, the DWb provides tools to empower the SMEs to express this knowledge.
DWb has many data exploration features, such as sort, filter, visual analytics, statistical analysis, cluster analysis, etc.
Additionally, we developed the adjudication plugin to support an expert-in-the-loop activity that captures a SME initial evaluation of a given material.
The SME should go through every material and answer a single question with few answer choices (for example: ``Does this candidate merit further exploration?'', with answers ``No'', ``Uncertain'', and ``Yes'') by evaluating only its constitution -- substructures, functional groups, ionic core.

The triage and adjudication processes are intertwined, such as the expert input during adjudication feeds back to the triage model (as highlighted by the blue arrow in Figure~\ref{fig:dwb-pag-workflows}).
By performing the adjudication, the SME is informing the triage AI models about the boundaries of the acceptable creative space and this feedback can be used to train a new triage model.

After a relatively small amount of labeling by SME (about 1000 can be evaluated in under an hour), the triage model is trained to identify the patterns of SME decisions and can rank the structures on its own.
In our initial experiments, after the triage step we were able to reduce to less than 10\% the dataset that will be made available for the next activity, the risk assessment.

%% file: tex/risk-assessment.tex
\subsection{Risk Assessment in Materials Discovery} \label{RiskAssessment}
In the discovery process of a novel material, evaluating the multiple dimensions that may prevent the material from succeeding for a purpose is essential. 
For example, some molecule candidates generated by AI models may not be successfully synthesized or represent a great danger to the environment due to their toxicity. 
In this work, we refer to \textit{risk assessment} as the activity of assessing the \textit{Probability-of-Success} (POS) of a novel material obtained from synthesizing a molecule candidate to fulfill its purpose. Evaluating the multiple dimensions (or risk factors) heavily depends on experts' knowledge from different areas. 
Computational tools, such as molecular dynamics simulators, may support the experts in this task. 
Despite the complexity and the heavy computational demand of those tools, ultimately, the SMEs have the final word due to the potential limitation and uncertainty of simulation results. 
These resource requirements (computational and human) limit the assessment of numerous candidates. 
So, the previous phases of triage and adjudication are fundamental to filtering the most promising ones.

The strong dependency on the tacit knowledge of multiple experts and the inherent uncertainty in the available supporting data characterizes the risk assessment in materials discovery as a knowledge-intensive process~\cite{kip}. 
Assessing data availability, quantity, and quality in this context is essential. 
We represent this assessment through the so-called \textit{Level-of-Knowledge} (LOK) metric~\citep{paltrinieri2019learning}. 
Methods that don't correctly handle all this complexity often lead to biased and inconsistent assessments. 
We apply the Knowledge-augmented Risk Assessment (KaRA) framework to tackle this challenge. 

\subsubsection{Knowledge-augmented Risk Assessment (KaRA) Framework} \label{kara}

The KaRA framework combines multiple AI techniques that consider SMEs' feedback on top of a structured domain knowledge base to support the risk assessment processes of candidates in knowledge-intensive contexts.
This framework is a generalization and extension of our previous work on geological risk assessment of petroleum system prospects~\cite{gra-eage,gsa-eage}.
The recent advances in AI and Knowledge Engineering provided the basis for the development of KaRA. 
Knowledge engineering practices and technologies are applied to represent and integrate the domain knowledge from multiple data sources and stakeholders and provide easy access to this knowledge through the DWb platform. 
At the same time, AI provides the appropriate tools for inference, prediction, uncertainty reduction, consensus reaching, etc. 

The methodology implemented by KaRA divides the assessment into multiple risk factors. 
Individual risk factor evaluation encompasses three phases: risk factor characterization, LOK assessment, and POS assessment.
Different AI agents support expert decisions at each one of those phases. As mentioned, the domain knowledge is structured and integrated by the KB managed by DWb. The KB also handles all the data used by KaRA components. KaRA aims to enable a co-creation environment for SMEs and AI agents to support the risk assessment process. 
Figure \ref{fig-kara-overview} summarizes the architecture, the methodology workflow, and some of the main features of KaRA. 
We discuss those features in the sequence according to each phase.

\input{fig/kara-overview}

\paragraph{Risk Factor Characterization} \label{kara-characterization}

We adopted a similar approach to \textit{Milkov}~\cite{milkov2015risk} and \textit{Jan-Erik et al.}~\cite{jan2000ccop} in the risk factor characterization phase.
For each risk factor, experts should answer a standard questionnaire showcasing the data analysis workflow results that could influence LOK or POS assessments. 
The characterization questionnaires structure the candidates in the KB, providing an easy way to represent them for supervised and unsupervised machine learning algorithms. 
One representative example used in the following phases is retrieving similar candidates in terms of characterization for comparison purposes. 
The differentiator of KaRA will be the ability to query and reason through the KB to retrieve and rank relevant evidence that may help the experts answer the characterization questions. 
The expert will curate the evidence recovered from the KB, keeping only the ones relevant to the answer. 
This feedback is used to retrain the ranking algorithms used in the evidence retrieval, which are continuously improved. 
This mechanism establishes a co-creation process where AI and experts collaborate to accurately characterize the candidate risk factors taking advantage of available evidence stored in the KB and the expert's tacit knowledge.

\paragraph{LOK Assessment} \label{kara-lok-assessment}

The role of the \textit{Level-of-Knowledge} (LOK) assessment phase is to establish fair and consistent LOK score scales dependent on the risk factor characterizations produced in the previous step to support the POS assessment process, such as done in \textit{Lowry et al.}~\cite{lowry2005advances} and \textit{Rose}~\cite{rose2001risk}. 
The LOK scales are created based on pairwise comparisons made by the experts and previously assessed, and peer-reviewed candidates present in KB. 
These candidates serve as training examples for a supervised machine learning algorithm that gives an initial LOK estimate, the so-called reference LOK. 
Experts' pairwise LOK comparisons calibrate this estimate by solving a Linear Programming (LP) model that determines the smallest overall adjustments in the LOK values needed to respect the comparisons. 
A rule-based inference strategy maintains the consistency of comparisons for a particular individual. 
The comparisons at the individual level capture the ranking opinions of each expert. 
The reference LOK estimates calibrated by these comparisons create the expert LOK scale. 
The system also determines the subset of consistent LOK comparisons that best represent experts' consensus by solving an Integer Programming (IP) model that minimizes the conflicts between the output comparisons and the overall expert opinions.
We obtain the so-called global LOK scale when these comparisons calibrate the reference LOK estimates with the same LP model previously described. 
Additional candidates and comparisons continuously adjust expert and global LOK scales. 
This approach is another example of a human-AI co-creation strategy applied to support an essential step in the discovery process.

\paragraph{POS Assessment} \label{kara-pos-assessment}
In the last phase, experts should collaborate to give a final POS value to the risk factor of the target candidate. 
The process collects individual POS assessments of multiple experts, and then a peer-review evaluation is conducted to reach a consensus value. 
Both steps constrain the experts' possible POS assessment values given a certain LOK level. 
The specification of how the LOK constrains the POS values is part of the specialization of the framework. 
KaRA's default method uses the so-called confidence-likelihood plot, a strategy devised in the Oil \& Gas industry~\cite{rose2001risk,lowry2005advances}. 
In this approach, POS values are constrained to the middle of the scale (around fifty percent) when the LOK is low, while at high LOK levels, the POS value should be either very low (around zero percent) or very high (around a hundred percent). 
The rationale is that in a low confidence situation, one should not be conclusive about the POS value, while in high confidence cases, one must be very assertive. 
The plots shown in Figure~\ref{fig:hornplots} specify this relationship by delimiting a region (in white) of allowed LOK (y-axis) and POS (x-axis) assessment pairs. 

\input{fig/pos-assessment}

Figure \ref{fig:expert-pos-assessment} shows the supporting plot for the individual expert's POS assessment step, while Figure \ref{fig:peer-review-pos-assessment} presents the one for the peer-review POS assessment. 
In Figure~\ref{fig:expert-pos-assessment}, we use the expert's LOK score to position the line on the y-axis. 
The expert is then constrained to give a POS value in the intersection between this line and the white region. 
The blue squares represent the assessments of similar candidates, where the blueness is proportional to the similarity with the target candidate. 
They serve as a reference to possibly maintain consistency with prior assessments. 
In Figure~\ref{fig:peer-review-pos-assessment}, the global LOK score determines the position of the LOK line, and the circles represent the multiple assessments of the different experts. 
With both information, the peer-review step may reach a final consensus driven by the multiple experts' opinions, potentially decreasing the biases in the process or at least making it more explicit.

At the end of the process, the evaluated candidates will have a detailed assessment of each dimension of interest (risk factors).
The generative models may use these results to calibrate their loss functions to increase the probability of generating new candidates with higher potential POS for each dimension. 
The complete workflow of the proposed materials discovery methodology establishes a continuous learning process through the collaboration and orchestration of AI agents and SMEs.

%% file: fig/kara-overview.tex
\begin{figure}[ht]
	\centering
	\includegraphics[width=0.9\textwidth]{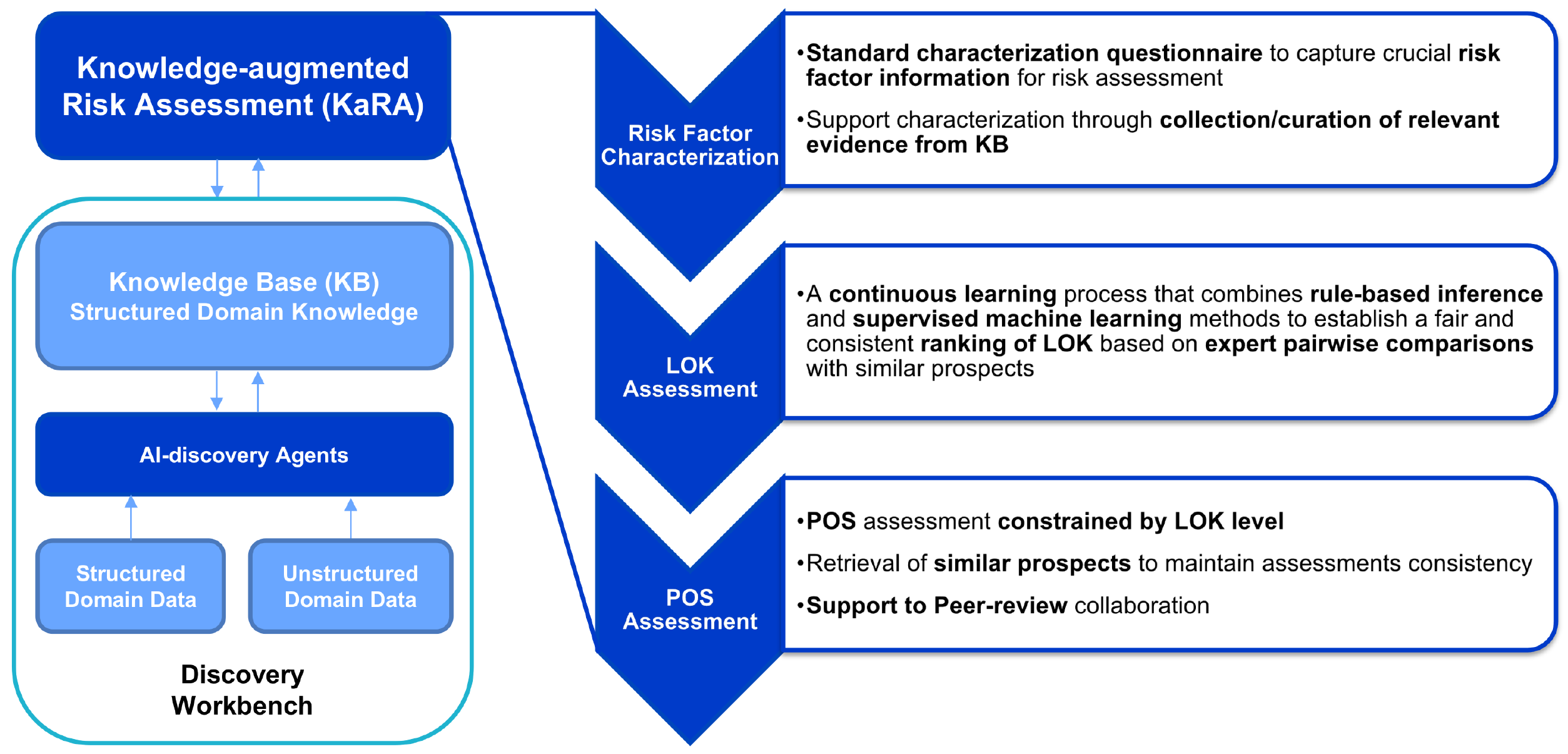}
	\caption{Overview of KaRA – methodology workflow and features.}
	\label{fig-kara-overview}
\end{figure}

%% file: fig/pos-assessment.tex
\begin{figure}
    \centering
    \begin{subfigure}{0.49\linewidth}
        \includegraphics[width=\linewidth]{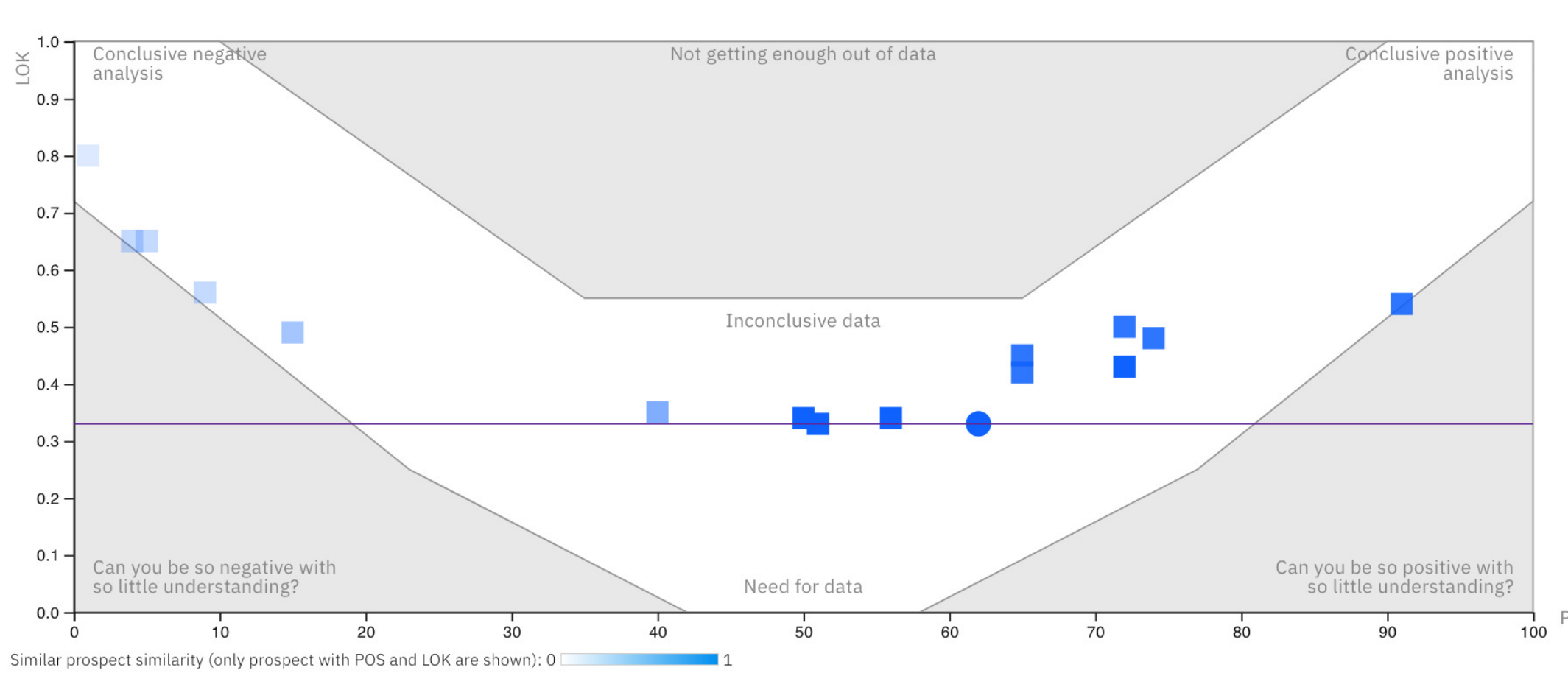} 
        \caption{Expert POS assessment}
        \label{fig:expert-pos-assessment}
    \end{subfigure}\hfill
    \begin{subfigure}{0.49\linewidth}
        \includegraphics[width=\linewidth]{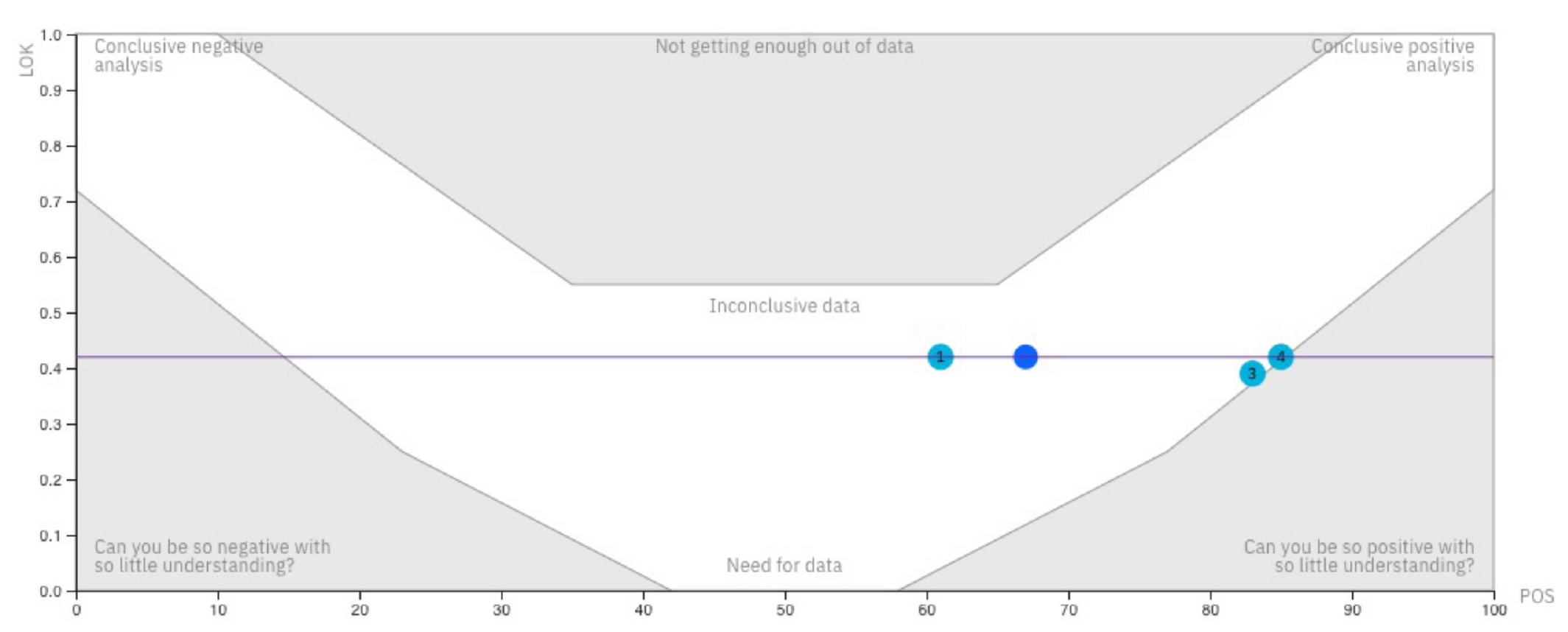}
            \caption{Peer-review POS assessment}
        \label{fig:peer-review-pos-assessment}
    \end{subfigure}
    \caption{Confidence-likelihood plots.}
    \label{fig:hornplots}
\end{figure}

%% file: tex/final-remarks.tex
\section{Final Remarks}
\label{sec:conclusion}

In this paper we discussed some of the challenges and opportunities faced when we try to bring AI to support an end-to-end discovery process.
We also introduced our ongoing work to create a method and its supporting technology that enable an efficient and effective collaboration between the SMEs and AI tools throughout the discovery process, applying a Human-AI co-creation approach.
The proposed approach recognizes the role of the SMEs as the main owner and decision-maker of the discovery process, aiming at providing AI-based technology to assist the SMEs in their abductive reasoning and to allow the capture of their knowledge for further reuse.

To develop and evaluate this approach, we have been working on a software framework called \textit{Discovery Workbench} (DWb) that enables the construction of end-to-end discovery applications.
DWb offers mechanisms to integrate multiple discovery tools and to support common discovery workflows by means of an integrated and user-friendly web-based interface.
DWb provides systematization to support experts with the discovery process and tracks the knowledge that is created along that process throughout its entire life cycle, from hypothesis generation and collection (\textit{i.e.} material candidates), to hypothesis testing (\textit{i.e.} automated simulations and lab experiments), and then to drawing conclusions (\textit{i.e.} visualizations and comparisons).

We have already created a PAG discovery application based on DWb, with promising initial results in terms of acceleration and systematization of the discovery process, mainly due to more flexible system integration and better knowledge management.
We have also started to explore other use cases of materials discovery to test the generality of the approach, such as meta-organic frameworks and biomass materials.
Despite the positive initial results, there are several areas that require further investigation, such as:
technologies to improve the incremental process of creating and curating an application-specific knowledge base for material discovery; more general design principles and implementation mechanisms to support the creation of more symbiotic interaction protocols between SMEs and AI peers, exploiting and advancing AI Explainability and Machine Teaching techniques;
multi-agent approaches to enable AI peers to orchestrate or even optimize the execution of the discovery process, requiring less SME intervention;
and the definition of qualitative and quantitative methodologies to assess the impact of the proposed approach and its supporting technologies on the discovery process.